\documentclass[runningheads]{llncs}
\usepackage{amsfonts,amsmath}
\usepackage{graphicx}
\usepackage{dsfont}
\usepackage{xcolor}
\usepackage{booktabs}
\usepackage{algorithm,algpseudocode}
\usepackage{subcaption}
\usepackage{multicol}
\usepackage{lipsum}
\usepackage{multirow}
\usepackage{gensymb}
\usepackage{pdfpages}
\usepackage[flushleft]{threeparttable}
\usepackage[justification=centering]{caption}

\begin{document}
\title{DeepLOC: Deep Learning-based Bone Pathology Localization and Classification in Wrist X-ray Images}
\titlerunning{DeepLOC}
\author{R. Dibo\inst{1}, A. Galichin\inst{1}, P. Astashev\inst{2}, D. Dylov\inst{1} and O.Y. Rogov\inst{1}}

%
\authorrunning{R. Dibo et al.}
%
\institute{Skolkovo Institute of Science and Technology, Moscow, Russia \and
Pirogov National Medical and Surgical Center, Moscow, Russia
\email{o.rogov@skoltech.ru}
}


%
\maketitle              
\begin{abstract}
In recent years, computer-aided diagnosis systems have shown great potential in assisting radiologists with accurate and efficient medical image analysis. This paper presents a novel approach for bone pathology localization and classification in wrist X-ray images using a combination of YOLO (You Only Look Once) and the Shifted Window Transformer (Swin) with a newly proposed block. The proposed methodology addresses two critical challenges in wrist X-ray analysis: accurate localization of bone pathologies and precise classification of abnormalities. The YOLO framework is employed to detect and localize bone pathologies, leveraging its real-time object detection capabilities. Additionally, the Swin, a transformer-based module, is utilized to extract contextual information from the localized regions of interest (ROIs) for accurate classification.

\keywords{Pathology localization \and transformers \and medical imaging}
\end{abstract}
\section{Introduction}

The wrist bones are important as they allow for a wide range of movements in the hand and wrist, and they are necessary for everyday activities such as grasping objects, writing, typing, and using tools\cite{rainbow2016functional}. Additionally, the wrist bones are important for transmitting forces from the forearm to the hand and fingers, which is necessary for activities such as lifting and carrying objects \cite{berger1996anatomy}. Injuries or conditions affecting the wrist bones can greatly impact the function of the hand and wrist, making it important to maintain their health and integrity.

Bone fractures are a significant global public health concern that greatly affects people's well-being and quality of life. In pediatric patients, they are the leading cause of wrist trauma \cite{hedstrom2010epidemiology,randsborg2013fractures}.According to \cite{mounts2011most}, fractures of the wrists were found to have the highest frequency of missed diagnoses, accounting for 32.4\% of cases. The manual analysis of X-ray images is the current method employed for fracture detection, but it is time-consuming, requiring experienced radiologists to review and report their findings to clinicians. Hence, a deficiency in the number of radiologists could significantly impact the timely provision of patient care \cite{rimmer2017radiologist}.
Extensive experiments are conducted on the modified GRAZPEDWRI-DX\cite{nagy2022pediatric} dataset of wrist X-ray images, demonstrating the effectiveness of the proposed approach. The results indicate significant improvements in both bone pathology localization and classification accuracy compared to state-of-the-art methods.


This paper encompasses the development of an advanced model capable of accurately detecting various bone diseases in X-ray images by employing bounding boxes for localization and subsequent disease classification. Unlike previous research that primarily concentrated on fractures as a classification problem, our model extends its scope to encompass a wider range of bone pathologies. To enhance the classification performance, a new approach incorporating multi-scale feature fusion and attention mechanisms is introduced, effectively capturing both local and global contextual information, enabling comprehensive representation learning for improved classification accuracy. Through the utilization of the proposed model, significant improvements in both speed and efficiency of the detection process have been achieved. 

The contributions of this research exposition can be summarized as follows. \textit{(1) Improved  architecture for medical vision using GAM attention}: we enhance the YOLOv7 architecture through the integration the GAM attention mechanism. The study identifies that incorporating the attention mechanism before the detection heads yields the best results in bone pathology localization. \textit{(2) Swin Transformer Integration:} We also explore the integration of a Swin Transformer along with the GAM mechanism, leading to further improvements in model performance. By identifying the optimal positions for incorporating these components before the detection heads, the study achieves favorable outcomes, advancing the state-of-the-art in bone pathology detection and recognition surpassing the current state-of-the-art results by 6\%.

\section{Related work}
\label{s:RW}


Bone fracture detection and classification using DL techniques has been developed during the last few years on both open source and clinical bone image datasets gathered from various medical devices.

Pathare et al. used preprocessing techniques to reduce the noise of the X-ray images; the median filter is used to specifically remove the salt and pepper noise. Edge detection methods are then used to find the edge in the images. Finally, segmentation methods that used the Hough transform were applied to analyze the edges and detect the fracture. The accuracy of this system was 75\%, and it was able to detect the main fractures only \cite{pathare2020detection}.

A deep neural network model was developed to classify healthy and fractured bones \cite{yadav2020bone}. Data augmentation techniques were used to address overfitting caused by the small dataset. The model employed a convolutional neural network architecture, with three experiments testing its performance. The proposed model achieved an accuracy of 92.44\% through 5-fold cross-validation. However, further validation on a larger dataset is needed to confirm its performance.

A set of 10 convolutional neural networks was developed to detect fractures in radiographs \cite{jones2020assessment}. Each network had variations in architecture and output structure but used the Dilated Residual Network\cite{Dilated} as the basis. They were individually optimized on a training set to predict the probability of a radiograph containing a fracture and, for some networks, the probability of each pixel being part of a fracture site. The ensemble output was obtained by averaging the individual network outputs. The dataset used for training consisted of 16 anatomical regions, with some regions over-represented due to imbalanced data.

In \cite{nguyen2021deep}, the authors proposed a deep learning-based method using YOLACT++ and the CLAHE algorithm for fracture detection in X-ray images of arm bones and trained YOLOv4 on a small dataset, incorporating data augmentation techniques to enhance the detector's performance. The method achieved an AP result of 81.91\%. However, the detection of fractures in more complex wrist bones was not addressed. Similarly, the study of \cite{hardalacc2022fracture} aimed to detect wrist fracture areas in X-ray images using deep learning-based object detection models with different backbone networks. The study also featured five ensemble models to develop a unique detection system that achieved an average precision of 0.864, surpassing the individual models. However, the study focused solely on fractures, and the final models were complex.

\textit{You Only Look Once (YOLOv7)} model is considered one of the fastest and most accurate real-time object detection model for computer vision tasks comparing to other YOLO versions and other object detection models \cite{wang2021scaled,bochkovskiy2020yolov4,glenn2022yolov5}. YOLO directly predicts the bounding boxes and class probabilities from the entire image in a single feedforward pass of the neural network. Therefore, it is considered as a one-stage detection model. As opposed to two-stage models as Faster R-CNN \cite{girshick2015fast}, which first propose regions of interest (RoIs) in the image before classifying and refining them. In a YOLO model, image frames are featured through a backbone, which is then combined and MIXed in the neck.

\textit{The Transformer} architecture enables information exchange among all entity pairs, such as regions or patches in computer vision. It has achieved remarkable performance in tasks like machine translation\cite{xiao-etal-2019-lattice}, object detection\cite{He_2022_CVPR}, multimodality reasoning\cite{NEURIPS2022_11332b6b}, image classification\cite{Lanchantin_2021_CVPR} and image generation\cite{razzhigaev}. However, the scalability of the Transformer poses challenges due to its quadratic complexity concerning the length of input sequences. To tackle this computation bottleneck, we suggested to use Swin transformer \cite{liu2021swin} which utilize non-overlapping local windows for self-attention computation while enabling cross-window consections. This way, the shifted window technique enhances the efficiency.

\section{Methods.}

\subsubsection{Dataset.}
\label{s:data}
The GRAZPEDWRI-DX dataset was used in this research\cite{nagy2022pediatric}. It is open source with 20327 annotated pediatric trauma wrist radiograph images of 6091 patients, treated at the Department for Pediatric Surgery of the University Hospital Graz. Several pediatric radiologists annotated the images by placing bounding boxes to mark nine different classes, which are: bone anomaly, bone lesion, foreign body, fracture, metal, periosteal reaction, pronator sign, soft-tissue and text. Each X-ray image can be with more than one class. After discussion with radiologists, it was suggested to mix the classes (''foreign body`` and ''metal`` were treated as one class name ''Foreign body``, ''Bone anomaly`` and ''Bone lesion`` were treated as one class called ''Bone lesion``) and to keep fracture and periosteal reaction classes which were considered the most important classes for detection. 
\subsubsection{YOLOv7. }
Object detector models feature a "neck" consisting of additional layers inserted between the backbone and head. 
The backbone is crucial for the model's accuracy and speed, composed of multiple convolutional layers that extract features at various scales. These features are then processed in the head and neck before being used to predict object locations and classes. The ELAN and ELAN-H modules, employing the CBS base convolution module, are significant parts of the Backbone component, responsible for feature extraction. In YOLOv7 architecture, MP (Max-Pooling), UP (Upsample), and RepConv (Reparameterized Convolution) are essential for the model's performance. MP downsamples feature maps to increase the receptive field, while UP upsamples to recover spatial information for better object localization. RepConv combines depthwise separable convolution and pointwise convolution to enhance model efficiency without sacrificing performance. These components enable YOLOv7 to detect objects accurately and efficiently in real-time applications. The model also incorporates the SPPCSPC module, a CSPNet with an additional SPP block that utilizes multiple pooling layers to extract region features at different scales. The inclusion of CSPNet reduces computational complexity, improves memory efficiency, and enhances inference speed and accuracy by addressing redundant gradient information.

\subsubsection{Transformer.}



For processing the patch tokens, multiple Swin Transformer blocks are employed, consisting of a shifted window-based multi-head self-attention module (MSA) and two layers of MLP, are employed. Layer normalization (LN) is applied before each MSA and MLP, with residual connections facilitating information flow. For a hierarchical representation, the number of tokens is progressively reduced using patch merging layers as the network deepens. The first patch merging layer concatenates the features of neighboring patches grouped in $2\times 2$ grids. A linear layer is then applied to the concatenated features, which are $4C$-dimensional features. This downsamples the number of tokens by a factor of $2\times 2=4$ (equivalent to a $2\times$ reduction in resolution), and the output dimension is set to $2C$. Then, Swin Transformer blocks are applied to transform features while maintaining a resolution of $\frac{H}{8} \times \frac{W}{8}$. This process is repeated for subsequent stages and resulting in a hierarchical representation of the feature maps.



\begin{figure}[t]
  \centering
  \includegraphics[width=1\textwidth]{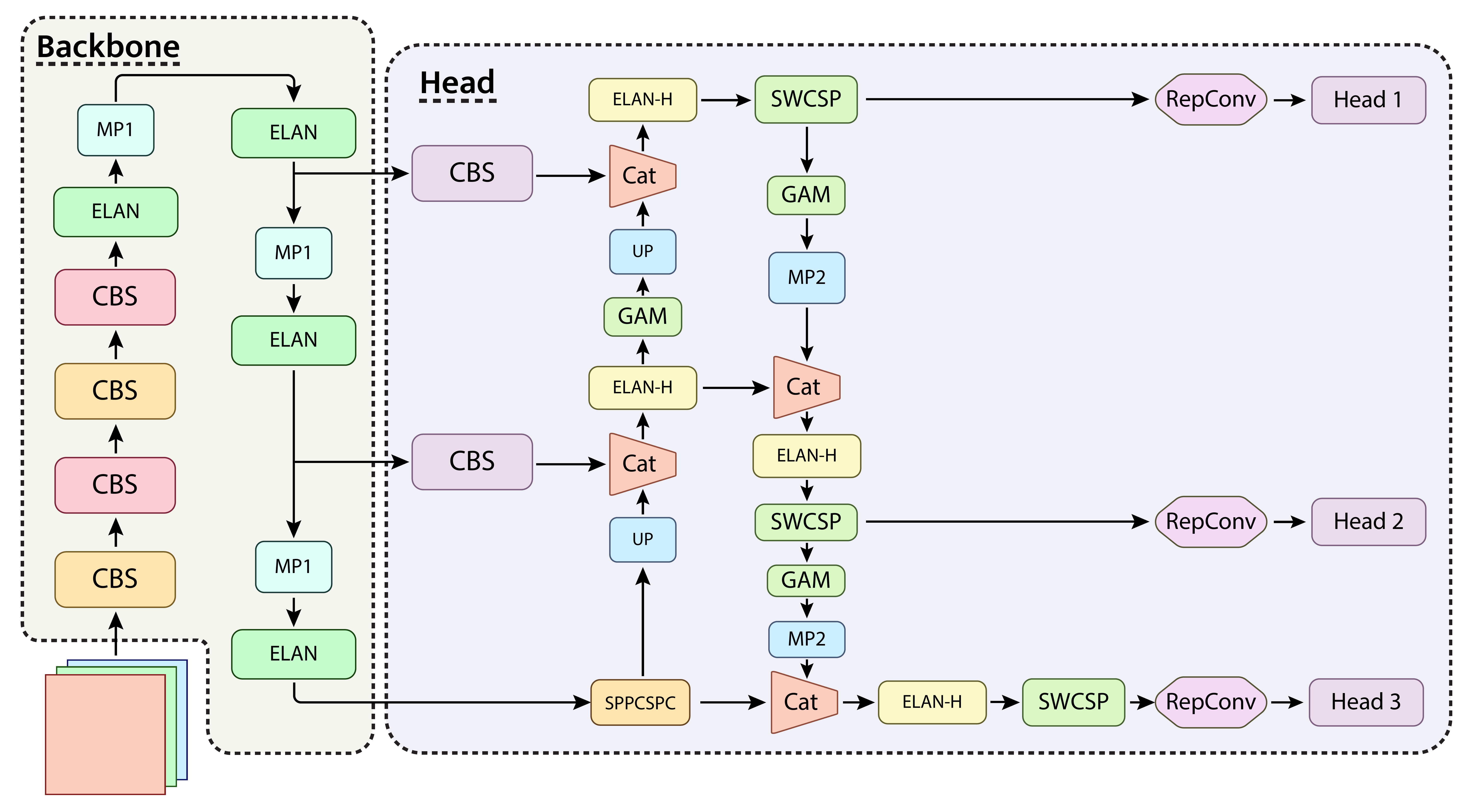}
  \caption{The proposed architecture with the GAM attention block and Swin transformer.}
  \label{fig:arch_final}
\end{figure}
Different attention processes that employ either channel attention or spatial attention were recently proposed. The convolutional block attention module (CBAM) considered both spatial and channel attention sequentially \cite{woo2018cbam}, but it ignored the channel-spatial interaction and as a result lose the cross-dimension information. Therefore, GAM was proposed \cite{liu2021global} where 3D permutation before channel attention was applied to preserve information across 3 dimensions and a feed-forward network which consists of two linear transformations with a ReLU activation in between to amplify cross-dimension channel-spatial dependencies. Then two convolutional layers for spatial information fusion were used with batch normalization and Relu in between.

\subsubsection{Proposed approach.} Based on the previous explanation for each block and method that will be used in the final architecture, the final model for bone pathology detection is denoted as DeepLOC (\textbf{Deep} \textbf{L}earning-based Bone Pathology \textbf{LO}calization and \textbf{C}lassification), which is obtained by inserting Swin transformer block (SWCSP) along with GAM attention block before each detection head of YOLOv7 architecture, and it is illustrated in Figure \ref{fig:arch_final} and the pseudocode for training and computing the loss is illustrated in Algorithm \ref{alg:loss}. Experimental evidence in Section \ref{s:experiments} indicates that the best placement of Swin and GAM blocks is the head part before the RepConv. Anchor-
based models employ a Feature-Pyramid-Network (FPN) which is presented in \cite{lin2017feature} to assist with the detection of objects of different scales. YOLOv7 also uses FPN to find the candidate bounding boxes.
Center Prior consists of anchor boxes whose anchor is sufficiently close to the target box center and whose dimensions closely align with the target box size.
\subsubsection{Loss function.} Following the work of\cite{CIOU}, we utilize the complete IoU metric and the penalty term for CIoU as $\mathcal{R}_{C I o U}=\frac{\rho^2\left(\mathbf{b}, \mathbf{b}^{g t}\right)}{c^2}+\alpha v$. Where the parameter $\alpha$ represents a positive trade-off and is defined as: $\alpha=\frac{v}{(1-I o U)+v}$. Therefore, it assigns greater importance to the overlap area factor for the regression, particularly when dealing with non-overlapping cases. The loss function can be formally defined as follows $\mathcal{L}_{C I o U}=1-I o U+ {R}_{C I o U}$.  To regulate the intensity of the edges $\lambda$ parameter is used\cite{wang2022yolov7}.

\begin{algorithm}[H]
\caption{DeepLOC loss calculation}
\label{alg:loss}
\begin{algorithmic}[1]
    \For{each head of FPN}
    \State find Center Prior anchor boxes
            \If{there is a matched target "ground truth"}
                \State calculate objectness\_loss $\mathcal{L}_{\text {O}}$ score as:
                
                    \quad \quad \textbf{\textit{BCE}} (predicted objectness probability, CIoU with the matched target)

            \Else { $\mathcal{L}_{\text {O}} = 0$}
            \EndIf
            
            \If{there are any selected anchor boxes candidates}
                \State calculate localization\_loss $\mathcal{L}_{\text {Loc}}$ score as:

                    \quad \quad \texttt{mean(1 - CIoU)} 
                    
                    \quad \quad between all candidate anchor boxes and their matched target

                \State calculate classification\_loss $\mathcal{L}_{\text {C}}$ score as:

                    \quad \quad \textbf{\textit{BCE}} (predicted class probabilities for each anchor box,
                    
                    \quad \quad one-hot encoded vector of the true class of the matched target)

            \Else \quad {$\mathcal{L}_{\text {Loc}} = \mathcal{L}_{\text {Loc}} = 0$}
            \EndIf
            
    \State Multiply $\mathcal{L}_{\text {O}}$ by corresponding FPN head weight "predefined hyperparameter"
    \EndFor
    \State Multiply each loss component: 
    
            $\mathcal{L}_{\text {O}}$, $\mathcal{L}_{\text {Loc}}$, $\mathcal{L}_{\text {C}}$ by their contribution weights (predefined ${\lambda}_{\text {O }}$, ${\lambda}_{\text {Loc }}$, ${\lambda}_{\text {C }}$).

    \State  Sum the already weighted loss components.
    
    $ \mathcal{L} _{\text {sample }}={\lambda}_{\text {Loc }}. \mathcal{L}_{\text {Loc}} + {\lambda}_{\text {O}}. \mathcal{L}_{\text {O}} +
        {\lambda}_{\text {C }}.\mathcal{L}_{\text {C}} $  
    
    \State Multiply the final loss value by the batch size.
    \State \textbf{return} $\mathcal{L}_{total}$

\end{algorithmic}
\end{algorithm}


\section{Experiments.}
\label{s:experiments}
All the models were trained from scratch using the GRAZPEDWRI-DX dataset. The different architectures were trained for 100 epochs with a batch size of 32. The image size was set to 640$\times$640, and the confidence threshold to \texttt{1e-3}. In the feature-extraction phase, the YOLO family networks were trained using the stochastic gradient descent (SGD) optimizer with a learning rate of \texttt{1e-2}, weight decay of \texttt{5e-4} and momentum of 0.937. All models were trained on an 4-Nvidia-V100s machine. Images were labeled with bounding box coordinates and class labels, which were then normalized for use in YOLOv7. The dataset was divided into 70\%, 20\%, and 10\% splits for training, validation, and testing respectively.


YOLOv7 was trained using data preprocessing techniques like CLAHE, Unsharp masking with median and Gaussian filters, and a proposed preprocessing method called MIX. The MIX method involves resizing images to 640$\times$640, applying mosaic and mixup with a 0.15 probability, rotation between -15$\degree$ and 15 $\degree$, and horizontal flipping with a 0.5 probability. Training was performed using the SGD optimizer and a batch size of 32. Performance was assessed with Average Precision (AP) computed at an Intersection over Union (IOU) of 0.5.

A total of nine anchor boxes are utilized in the training process. The width and height of each anchor box for each scale are illustrated in Table \ref{table:sizes}.
\vspace{-2em}

\begin{table}[]
\centering
\caption{The dimensions (width and height) of each anchor box for every scale/head in YOLOv7, which were determined based on the sizes of the bounding boxes in the annotations.}
\begin{tabular}{@{}cccc@{}}
\toprule
\multicolumn{1}{l}{Scale\textbackslash{}Size} &
  \begin{tabular}[c]{@{}c@{}}Anchor\_1\\ (Width, Height)\end{tabular} &
  \begin{tabular}[c]{@{}c@{}}Anchor\_2\\ (Width, Height)\end{tabular} &
  \begin{tabular}[c]{@{}c@{}}Anchor\_3\\ (Width, Height)\end{tabular} \\ \midrule
Small  & 12,16   & 19,36   & 40,28   \\
Medium & 36,75   & 76,55   & 72,146  \\
Large  & 142,110 & 192,243 & 459,401 \\ \bottomrule
\end{tabular}
\centering

\label{table:sizes}
\end{table}

Patch size for the first stage of Swin transformer was set to $4 \times 4 $, window\_size was 7, number of heads in MHSA is 4, The image size was set to 672 × 672 to match the used window\_size if Swin block is added.
The query dimension of each head is d = 32, and the expansion layer of each MLP is $\alpha$ = 4 with SilU activation\cite{silu}. CBAM and GAM attention mechanisms were tested while applied to different positions of the YOLOv7 architecture (backbone, neck, and before detection heads). The reduction ratio in MLP in the channel attention submodule which was set to 4. The same reduction ratio was used in the spatial attention submodule.

\section{Results and discussion.}
The results of the trained models are illustrated in Table \ref{table:prep}.

\begin{table}
      \centering
      \parbox{.45\linewidth}{
      \caption{The results of training YOLOv7 model with different preprocessing and augmentation techniques with SGD optimizer and batch size 32.}
      \label{}
      \begin{tabular}{cccccc}
        \toprule
        \multirow{2}{*}{Method} & \multicolumn{5}{c}{AP @0.5} \\  
         &
          \multicolumn{1}{l}{F} &
          \multicolumn{1}{l}{FB} &
          \multicolumn{1}{l}{PR} &
          \multicolumn{1}{l}{BL} &
          \multicolumn{1}{l}{mAP} \\ \midrule
        \begin{tabular}[c]{@{}c@{}}UM (M)\end{tabular} &
          \multicolumn{1}{c}{0.929} &
          \multicolumn{1}{c}{0.572} &
          \multicolumn{1}{c}{\textbf{0.648}} &
          \multicolumn{1}{c}{0.049} &
          0.550 \\ 
        \begin{tabular}[c]{@{}c@{}}UM (G)\end{tabular} &
          \multicolumn{1}{c}{0.932} &
          \multicolumn{1}{c}{0.891} &
          \multicolumn{1}{c}{0.605} &
          \multicolumn{1}{c}{0.04} &
          0.617 \\ 
        \begin{tabular}[c]{@{}c@{}}CLAHE\end{tabular} &
          \multicolumn{1}{c}{0.933} &
          \multicolumn{1}{c}{0.870} &
          \multicolumn{1}{c}{0.617} &
          \multicolumn{1}{c}{0.044} &
          0.621 \\ 
        MIX &
          \multicolumn{1}{c}{\textbf{0.938}} &
          \multicolumn{1}{c}{\textbf{0.912}} &
          \multicolumn{1}{c}{0.606} &
          \multicolumn{1}{c}{\textbf{0.057}} &
          \textbf{0.628} \\ \bottomrule
        \label{table:prep}
        
        \end{tabular}
        \centering
        \label{table:prep}
            \begin{tablenotes}
            \small
            \item \textbf{Note.} Here UM is the unsharp masking (UM), (M) stands for (using median filter), (G) for (Gaussian filter). WS is window size, and CL is clip limit for CLAHE. The classes are: fractures (F), foreign body (FB), Periosteal reaction (PR), Bone lesion (BL). mAP is calculated as total. CLAHE is used with parameters WS=8 and CL= 4.
            \end{tablenotes}
        }
        \qquad
        \parbox{.45\linewidth}{
        \caption{The results of training YOLOv7 model with CBAM and GAM attention mechanisms in different positions of the network with SGD optimizer and batch size 32 on the test dataset.}
        \begin{tabular}{@{}ccccc@{}}
        \toprule
        Model & \multicolumn{1}{l}{P} & \multicolumn{1}{l}{R} & \multicolumn{1}{l}{F1} & \multicolumn{1}{l}{mAP} \\ \midrule
        Yolov7                  & 0.556 & 0.582 & 0.569 & 0.628 \\
        GAM (bh 1)     & 0.745 & 0.574 & 0.646 & 0.634 \\
        GAM (bh 2)     & 0.743 & 0.543 & 0.627 & 0.636 \\
        GAM (bh 3)     & 0.749 & 0.614 & 0.675 & 0.638 \\
        GAM (ba)  & 0.751 & 0.621 & 0.680 & 0.638 \\
        CBAM (ba) & 0.709 & 0.593 & 0.646 & 0.633 \\ \bottomrule
        \end{tabular}
        \centering
        \label{table:att}
            \begin{tablenotes}
            \small
            \item \textbf{Note.} (bh1), (bh2), (bh3) stand for inserting before head 1, head 2 and head 3 respectively; (ba) stands for inserting before all detection heads.
    \end{tablenotes}
        }
\end{table}

Based on the obtained results, the ''MIX`` method, which demonstrated the highest mean average precision value, will be adopted as the preferred preprocessing and data augmentation technique for the subsequent sections.


we investigate the effects of inserting the GAM attention mechanism before each individual head and before all three heads collectively. The results of these experiments are presented and summarized in the subsequent Table \ref{table:att}. 



Additionally, we explore the consequences of incorporating the Swin Transformer in the backbone of the architecture and before the detection heads. Finally, we examine the effects of incorporating both the Swin Transformer and attention module in the YOLOv7 architecture, specifically before the detection heads. Our aim is to assess the impact of this combination on the overall performance of the system. The results of these experiments are presented and summarized in the following Table \ref{table:results}.

The proposed model DeepLOC yielded the most favorable outcomes, surpassing YOLOv7 alone by 2.6\% and outperforming the insertion of Swin blocks before the detection heads by 0.3\%. This model was chosen due to its superior average precision (AP) in the bone lesion class, exhibiting a notable improvement of 5.8\% compared to YOLOv7 alone. Moreover, it exhibited a 2.6\% enhancement in the foreign body class. Figure 2 displays precision-recall curves obtained during the testing process.
\begin{figure}
  \begin{minipage}[b]{.45\linewidth}
    \centering
    \label{fig:enter-label}
    \includegraphics[width=1\linewidth]{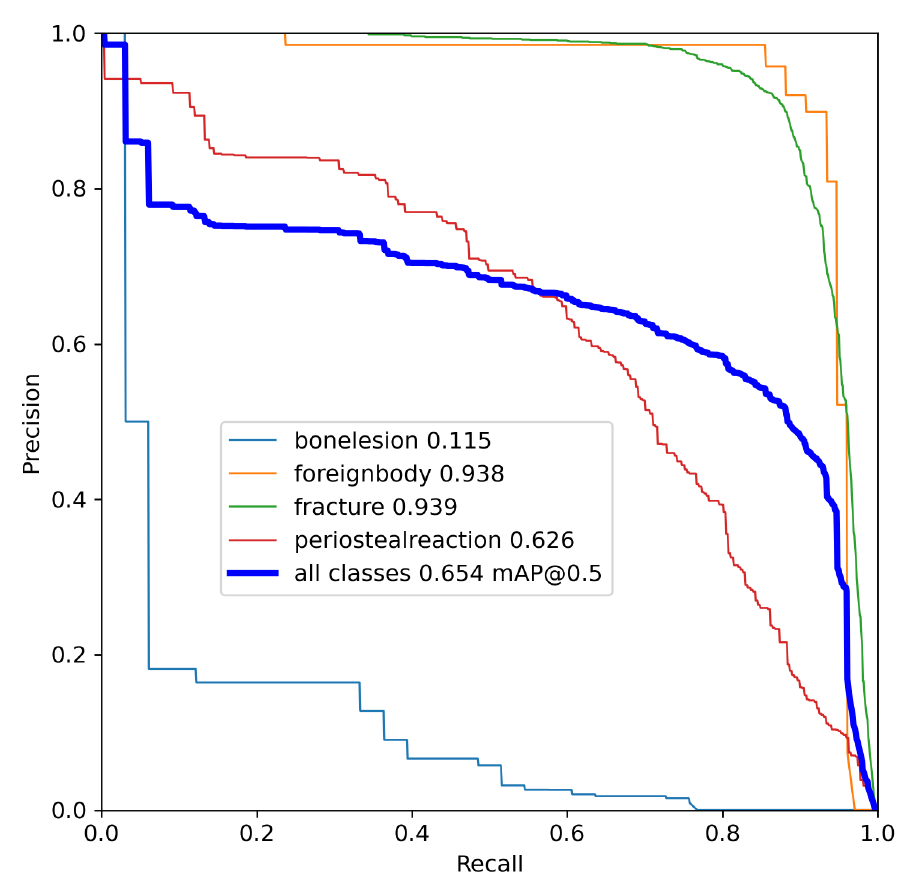}
    \captionof{figure}{Precision-Recall curves using DeepLOC model.}
  \end{minipage}
  \begin{minipage}[b]{.45\linewidth}
    \centering
    \label{t:metrics}
    \begin{tabular}{@{}lcccccc@{}}
    \toprule
     Class & Labels &  P &  R & mAP@.5 &  mAP@.5-0.95 \\
    \midrule
    All & 2161 & 0.64 & 0.632 & \textbf{0.654} & 0.398 \\
    BL  & 33 & 0.165 & 0.108 & 0.115 & 0.0418 \\
    FB  & 76 & 0.871 & 0.934 & 0.938 & 0.76 \\
    F  & 1735 & 0.877 & 0.889 & 0.939 & 0.535 \\
    PR & 317 & 0.648 & 0.599 & 0.626 & 0.257 \\ \bottomrule
    \end{tabular}%
    \captionof{table}{Precision, Recall, \text { mAP@.5 } and \text { mAP@.5-0.95 } for all classes using DeepLOC model.}
  \end{minipage}
\end{figure}



\begin{table}[]
\centering
\small
\caption{Performance comparison for the difference models.}
\begin{tabular}{@{}cccccc@{}}
\toprule
\multirow{2}{*}{Model}                                                           & 
\multicolumn{5}{c}{AP}                                           \\ 
 & Fracture & Foreign body  & Periosteal reaction & Bone lesion & mAP \\ 
 \midrule
Yolov7                                                                           & 0.938    & 0.912       & 0.606              & 0.057      & 0.628 \\ 
Yolov7+CBAM                                                                    & 0.932    & 0.922       & 0.608              & 0.07       & 0.633 \\ 
Yolov7+GAM                                                                     & 0.933    & 0.932       & 0.607              & 0.08       & 0.638 \\ 
\begin{tabular}[c]{@{}c@{}}Yolov7+SW(ba)\end{tabular}            & 0.936    & 0.928       & \textbf{0.636}     & 0.106      & 0.651 \\ 
\begin{tabular}[c]{@{}c@{}}Yolov7+SW(B+ba)\end{tabular} & 0.932    & 0.926       & 0.589              & 0.02       & 0.651 \\ 
\begin{tabular}[c]{@{}c@{}}Yolov7+SW+CBAM(ba)\end{tabular}       & 0.938    & 0.932       & 0.635              & 0.07       & 0.646 \\ 
\begin{tabular}[c]{@{}c@{}}\textbf{DeepLOC}\end{tabular} & \textbf{0.939} & \textbf{0.938} & 0.626 & \textbf{0.115} & \textbf{0.654} \\ \bottomrule
\end{tabular}
\centering
\label{table:results}
    \begin{tablenotes}
    \small
    \item \textbf{Note.} Here B stands for inserting the block in the Backbone, (ba) stands for inserting before all the detection heads. SW is the Swin transformer block. Reproduced with the official code under fine-tuning setting for fair comparison with SGD and batch  32.
    \end{tablenotes}
\end{table}

\begin{figure}
    \centering
    \includegraphics[width=0.8\textwidth]{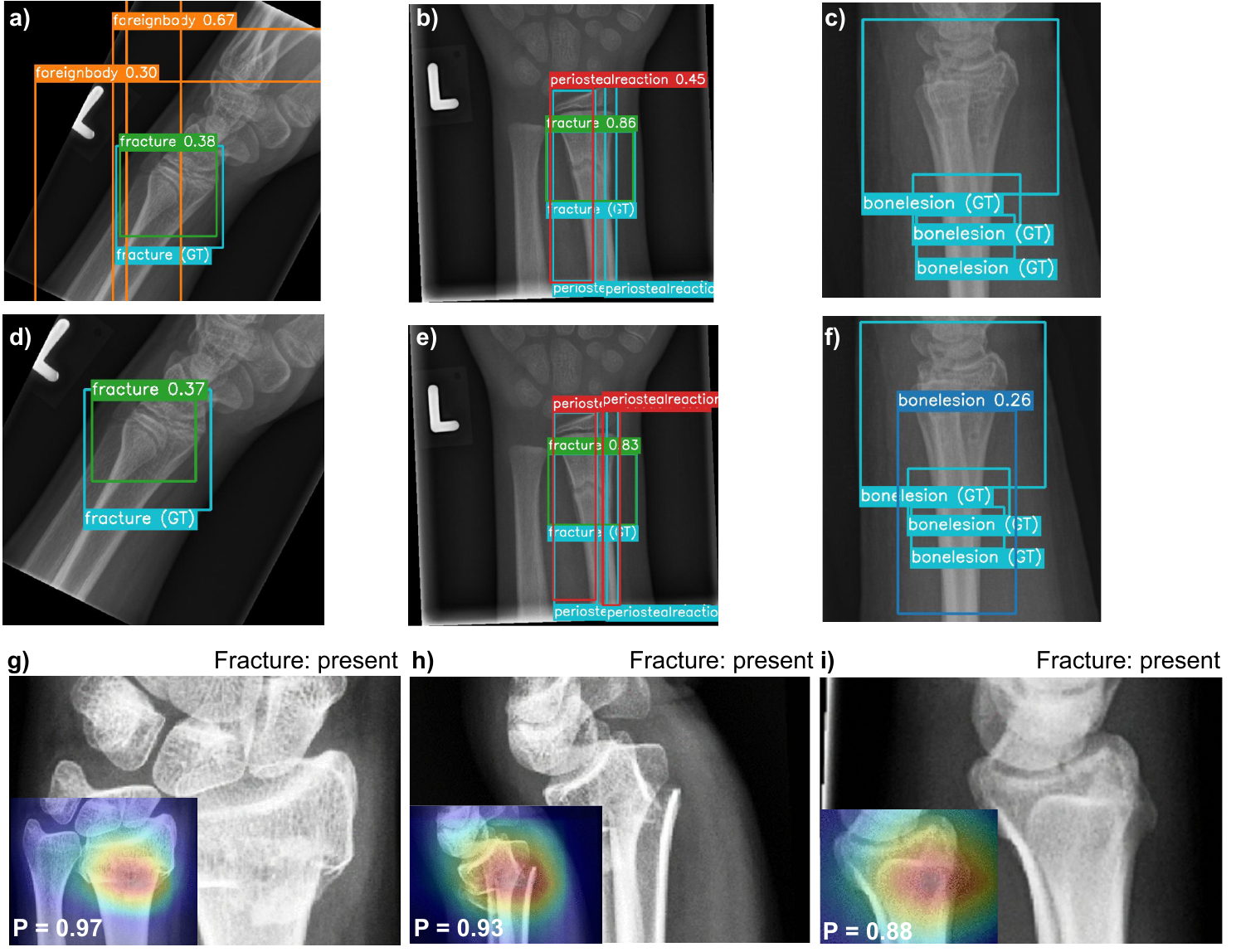}
    \caption{Original X-Ray images and feature localization heatmaps. Each sub-figure presents an input image with GradCAM++ heatmaps as insets with the prediction probability inside.}
    \label{fig:images}
\end{figure}

Furthermore, experiments were performed to compare various models based on parameters, GFLOPs, layers, inference time  (including pre-processing and post-processing time), and batch 32 inference time. The resulting comparisons are presented in the Table \ref{table:results2}, illustrating the outcomes of the analysis. In addition, Table 3 illustrates the detailed results in terms of precision, recall, mAP for all classes using the DeepLOC model. In Fig. \ref{fig:images}, with the ground truth represented by the blue bounding boxes, it can be observed that the YOLOv7 \textbf{(a)}-\textbf{(c)} architecture alone exhibited misclassifications where the model incorrectly identified the edges of the image as instances of the foreign body class. Additionally, the model accurately predicted only one out of the two instances of periosteal reaction present in the ground truth. Here YOLOv7 model fails to detect any instances of the bone lesion classes while DeepLOC correctly localizes the pathology (e.g., Fig. \ref{fig:images} \textbf{(d)}-\textbf{(f)}.We additionally visualized the GradCAM++\cite{GCPP} heatmaps in Figure \ref{fig:images}. Images \textbf{g)} and \textbf{h)} show examples of true positive cases from the test set with easily visible fractures while in \textbf{(i)} the pathology can hardly be seen with the naked eye. Remarkably, for the True Positive cases in the dataset, on subplots \textbf{(g)}-\textbf{(i)}, DeepLOC identified the correct areas, where the distal radius fractures are present. In contrast, when employing our proposed model DeepLOC, notable improvements are observed. The model successfully addresses the issue of erroneous edge prediction, accurately detects both instances of periosteal reaction, and demonstrates the ability to predict one out of three instances of the Bone lesion class.
\vspace{-2em}

\begin{table}[]
\centering
\caption{Comparison of different model performances.}
\begin{tabular}{@{}lccccc@{}}
\toprule
Model & Param(M) & GFLOPs & Layers & $t_{i}$ & Batch average time(ms) \\ \midrule
Yolov7+SW(B+ba) & 76.08 & 1614.6 & 449 & 26.7 & 15.8 \\
Yolov7+SW+CBAM(ba)& 42.04 & 317.8 & 431 & 20.3 & 5.5 \\
DeepLOC & 43 & 321.9 & 428 & 24.8 & 4.4 \\ \bottomrule
\end{tabular}%
\centering
\label{table:results2}
    \begin{tablenotes}
    \small
    \item \textbf{Note.} Here B stands for inserting the block in the Backbone, (ba) stands for inserting before all the detection heads. SW is Swin transformer block. $t_{i}$ is the inference time in ms.
    \end{tablenotes}
\end{table}
\vspace{-2em}

\section{Conclusion.}
We emphasize the potential of incorporating new transformer architectures, specifically designed for object detection tasks, to further enhance performance. The exploration of novel attention mechanisms and the refinement of training processes to address false positives and false negatives show promising results. Optimizing confidence thresholds dynamically, expanding the dataset, and conducting ablation studies to identify influential architectural elements are suggested as important areas for further research. These efforts aim to refine and optimize object detection models for pediatric wrist trauma X-ray images, ultimately advancing the state-of-the-art in this domain.

\section{Acknowledgements.}
The work was supported by Ministry of Science and Higher Education grant No. 075-10-2021-068.

\bibliographystyle{splncs04}
\bibliography{biblio}
\end{document}